# A Comparison of Lauritzen-Spiegelhalter, Hugin, and Shenoy-Shafer Architectures for Computing Marginals of Probability Distributions

**Vasilica Lepar**
Institute of Informatics
University of Fribourg
Site Regina Mundi, Rue Faucigny 2
CH-1700, Fribourg, Switzerland
vasilica.lepar@unifr.ch

**Prakash P. Shenoy**
School of Business
University of Kansas
Summerfield Hall
Lawrence, KS 66045-2003, USA
pshenoy@ukans.edu

## Abstract

In the last decade, several architectures have been proposed for exact computation of marginals using local computation. In this paper, we compare three architectures—Lauritzen-Spiegelhalter, Hugin, and Shenoy-Shafer—from the perspective of graphical structure for message propagation, message-passing scheme, computational efficiency, and storage efficiency.

## 1 INTRODUCTION

In the last decade, several architectures have been proposed in the uncertain reasoning literature for exact computation of marginals of multivariate discrete probability distributions. One of the pioneering architectures for computing marginals was proposed by Pearl [1986]. Pearl's architecture applies to singly connected Bayes nets. For multiply connected Bayes nets, Pearl [1986] proposed the method of conditioning to reduce a multiply connected Bayes net to several singly connected Bayes nets.

In 1988, Lauritzen and Spiegelhalter [1988] proposed an alternative architecture for computing marginals that applies to any Bayes net. Subsequently, Jensen et al. [1990a, b] proposed a modification of the Lauritzen-Spiegelhalter architecture. We call this architecture the Hugin architecture since this architecture is implemented in Hugin, a software tool developed by the same group. Recently, this architecture has been abstracted by Lauritzen and Jensen [1996] so that it applies more generally to other domains including the Dempster-Shafer's belief function theory.

Inspired by the work of Pearl, Shenoy and Shafer [1986] first adapted and generalized Pearl's architecture to the case of finding marginals of joint Dempster-Shafer belief functions in join trees. Later, motivated by the work of Lauritzen and Spiegelhalter [1988] for the case of probabilistic reasoning, they proposed an abstract framework for computing marginals in join trees that applies to any domain satisfying some axioms [Shenoy and Shafer 1990]. Recently, Shenoy [1997] has proposed a refinement of join trees, called binary join trees, designed to improve the computational efficiency of the Shenoy-Shafer architecture. We refer to this architecture as the Shenoy-Shafer architecture. In a sense, the Shenoy-Shafer architecture can be considered as an adaptation of Pearl's propagation scheme to the join tree graphical structure.

In this paper, we compare the Lauritzen-Spiegelhalter (LS), Hugin, and Shenoy-Shafer (SS) architectures from the perspective of graphical structure for message propagation, the message passing scheme, storage efficiency, and computational efficiency.

Our main findings are as follows. The Hugin architecture is more computationally efficient than the LS architecture, and less storage efficient than the LS architecture. This is not surprising. What is surprising is that on an average basis, the SS architecture is computationally more efficient than the Hugin architecture. It is commonly believed that the Hugin architecture is the most efficient architecture for computing marginals, but our experiments do not support this belief.

## 2 BAYESIAN NETWORKS

In this section, we will first define a Bayesian network probability model and then describe an example: Lauritzen and Spiegelhalter's [1988] Chest Clinic problem.

### 2.1 BAYESIAN NETWORKS

We denote variables by uppercase Roman letters, A, B, C, etc., and the set of all variables by $\Psi$. We denote subsets of variables by lowercase Roman alphabets c, s, t, etc. We denote the set of possible states of a variable X by $\Omega_X$, and we assume that the set of possible states of a subset c of variables is the Cartesian product of the state space of individual variables in the subset c, $\Omega_c = \times\{\Omega_X \mid X \in c\}$. We denote states of a subset of variables by lowercase boldfaced letters such as **x**, **y**, etc. If **x** is a state of c and b $\subseteq$ c, then $\mathbf{x}^{\downarrow b}$ denotes the projection of **x** to b obtained by simply dropping states of variables in c \ b. Of course, $\mathbf{x}^{\downarrow b} \in \Omega_b$.

Suppose c is a subset of variables. A *potential* for c is a function $\chi: \Omega_c \to \mathbb{R}^+$, where $\mathbb{R}^+$ is the set of non-negative real numbers. We call c the *domain* of potential $\chi$. We will denote potentials by lowercase Greek letters.

We define multiplication of potentials as follows. Suppose $\alpha$ is a potential for a, and suppose $\beta$ is a potential for b. Then $\alpha \otimes \beta$, read as $\alpha$ times $\beta$, is a potential for $a \cup b$ defined as follows: $(\alpha \otimes \beta)(\mathbf{x}) = \alpha(\mathbf{x}^{\downarrow a}) \beta(\mathbf{x}^{\downarrow b})$ for all $\mathbf{x} \in \Omega_{a \cup b}$.

We define marginalization of potentials as follows. Suppose $\alpha$ is a potential for a and suppose $b \subseteq a$. Then the *marginal* of $\alpha$ to b, denoted by $\alpha^{\downarrow b}$, is a potential for b defined as follows: $\alpha^{\downarrow b}(x) = \sum \{\alpha(x, y) \mid y \in \Omega_{a \setminus b}\}$ for all $x \in \Omega_b$.

A *Bayesian network* model consists of a connected acyclic digraph $G = (\Psi, \Delta)$, and a set of conditional potentials $\{\kappa_V\}_{V \in \Psi}$, where $\Psi$ represents the set of variables and $\Delta$ denotes the set of directed arcs between pairs of variables. An acyclic digraph is a finite oriented graph with no multiple arcs, and no directed cycles. If V and W are variables in $\Psi$ and there is a directed arc from W to V, written as $W \rightarrow V$, then we say V is a *child* of W, and W is a *parent* of V. Let $Pa(V) = \{W \in \Psi \mid W \rightarrow V\}$ denotes the set of *parents* of V. The conditional potentials $\{\kappa_V\}_{V \in \Psi}$ satisfy the following condition: $\kappa_V: \Omega_{\{V\} \cup Pa(V)} \rightarrow \mathbb{R}^+$ is such that $\kappa_V^{\downarrow Pa(V)}(x) = 1$ for every $x \in \Omega_{Pa(V)}$. The assumption underlying a Bayesian network model is that the prior joint probability distribution $P(\Psi)$ is given by $P(\Psi) = \otimes\{\kappa_V \mid V \in \Psi\}$ For a more detailed description of a Bayesian network model, see [Lauritzen and Spiegelhalter 1988, and Pearl 1986].

## 2.2 THE CHEST CLINIC PROBLEM

In this section, we will first describe Lauritzen and Spiegelhalter's [1988] hypothetical Chest Clinic problem, and next, a Bayesian network model for it.

> Shortness-of-breath (dyspnoea) may be due to tuberculosis, lung cancer, bronchitis, none of them, or more than one of them. A recent visit to Asia increases the chances of tuberculosis, while smoking is known to be a risk factor for both lung cancer and bronchitis. The results of a single chest X-ray do not discriminate between lung cancer and tuberculosis, as does neither the presence nor absence of dyspnoea.

**Figure 1.** A Bayesian Network for the Chest Clinic Problem

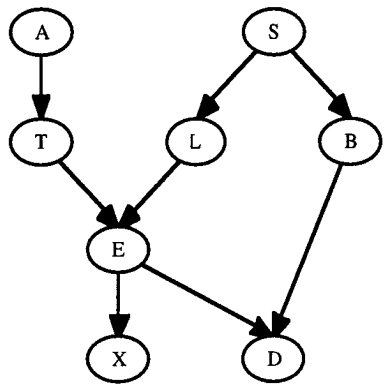

This problem is modeled as a Bayesian network as shown in Figure 1. In this network, A denotes the variable visit to Asia?, S denotes Smoker?, T denotes Has Tuberculosis?, L denotes Has Lung Cancer?, B denotes Has Bronchitis?, E denotes Has Either Tuberculosis or Lung Cancer, X denotes Has positive X-ray?, and D denotes Has dyspnoea?. We assume that all variables have binary state spaces. We will denote the conditional potential for a variable by its corresponding Greek letter. Thus, $\alpha$ will denote the prior probability distribution for A, $\tau$ will denote the conditional potential P(T | A), etc.

## 3 LAURITZEN-SPIEGELHALTER ARCHITECTURE

In this section, we describe the Lauritzen-Spiegelhalter architecture for computing marginals.

In a probabilistic model, we make inferences by computing the marginal of the joint probability distribution for the variables of interest. For simplicity, we will assume that we are interested in the marginal for all variables. When we have a large number of variables, computing the joint is computationally intractable. However, when the conditional potentials have small domains, we can compute the marginals of the joint without explicitly computing the joint.

In the LS architecture, first we construct a join tree called a junction tree, and then we propagate messages in the junction tree. A *join tree* is a tree whose nodes are subsets of variables such that if a variable is in two distinct nodes, then the variable must be in every node on the path between the two nodes. We will call the join tree whose nodes are the cliques of the triangulated moral graph a *junction tree*. This data structure enables local computations with potentials on domains within the cliques. A junction tree for the Chest Clinic problem is shown in Figure 2.

**Figure 2.** A Junction Tree for the Chest Clinic Problem

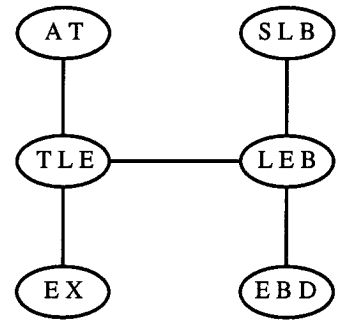

Next, we associate each conditional potential $\kappa_V$ with a smallest clique that contains the subset $\{V\} \cup Pa(V)$. If we have observations, we model these as potentials and associate the potentials with a smallest clique that includes the domain of the potential. If a clique has more than one potential associated with it, then we will assume that the product of these potentials is associated with the clique.

For the Chest Clinic problem, suppose we have evidence that the patient has visited Asia and has Dyspnoea. We model this evidence as potentials $o_A$ for $\{A\}$ and $o_D$ for $\{D\}$. It is easy to show that given the evidence, the posterior joint distribution is proportional to the product of all potentials including $o_A$ and $o_D$.

Next we pick a maximal node of the junction tree (that has the biggest state space) to be the root, and direct all edges of the junction tree toward the root. The propagation in Lauritzen and Spiegelhalter's architecture is done in two passes, inward and outward. In the inward pass, each node send a message to its inward neighbor, and in the outward pass, each node sends a message to its outward neighbors. Precise rules are as follows [Shafer 1996].

**Inward Pass** (see Figure 3):

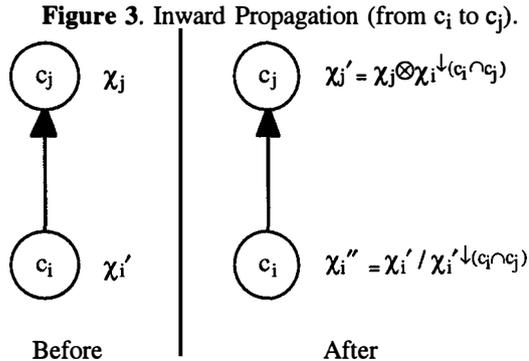

Figure 3. Inward Propagation (from $c_i$ to $c_j$).

Before | After

- **Rule 1**. Each node waits to send its message to its inward neighbor until it has received a message from all its outward neighbors. If a node has no outward neighbors, it can send a message right away.

- **Rule 2**. When a node is ready to send a message to its inward neighbor, it computes the message by marginalizing its current potential to its intersection with the inward neighbor. It sends this message to its inward neighbor, and then it divides its own current potential by the message.

- **Rule 3**. When a node receives a message from its outward neighbor, it replaces its current potential with the product of that potential and the message.

The inward pass ends when the root has received a message from all its outward neighbors.

Outward Pass (see Figure 4):

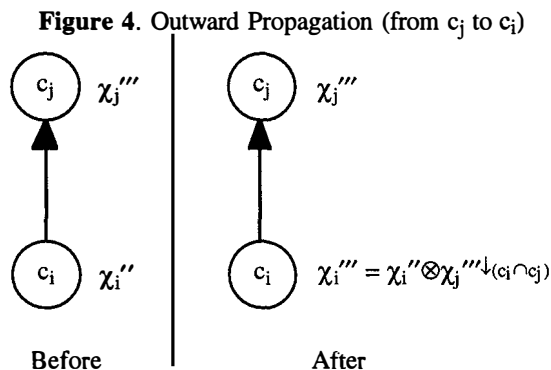

Figure 4. Outward Propagation (from $c_j$ to $c_i$)

Before | After

- **Rule 1**. Each node waits to send its messages to its outward neighbors until it has received the message from its inward neighbor. The root, which has no inward neighbor, can send a message right away.

- **Rule 2**. When a node is ready to send a message to its outward neighbor, it computes the message by marginalizing its current potential to its intersection with the outward neighbor. It sends this message to its outward neighbor.

- **Rule 3**. When a node receives a message from its inward neighbor, it replaces its current potential with the product of that potential and the message.

The outward pass ends when all leaves have received messages from their inward neighbors.

At the end of the outward pass, the potential associated with each clique is the marginal of the posterior joint for the clique (up to a normalization constant).

At the end of the outward pass, we have the marginal of the posterior distribution at each clique. However, the stated task is the computation of the marginal of the posterior for each variable in the Bayes net. We can compute the marginal for a variable from any clique marginal that contains the variable. Since it is more efficient to compute the marginal from a smaller clique, we will do so from a smallest clique that contains the variable. For example, to compute the marginal for E in the Chest Clinic problem, we can do so from the marginals of the following cliques: {T, L, E}, {L, E, B}, {E, B, D} and {E, X}. Since {E, X} is the clique with the smallest number of states, it is most efficient to compute the marginal for E from {E, X}. Of course, this strategy ignores the computational cost of identifying a smallest clique.

## 4   HUGIN ARCHITECTURE

In this section, we sketch the Hugin architecture. Although it was initially described for computing marginals of probability distributions [Jensen *et al.* 1990a, b], it has been recently extended by Lauritzen and Jensen [1996] so that it is more widely applicable to domains that satisfy some axioms.

We start by assuming that we have a junction tree and the corresponding probability potentials for each clique. We introduce a storage register between every two cliques whose domain is the intersection of the two cliques. We call this storage register a *separator*. Pick any node to be the root. The propagation in Hugin architecture is done in two passes, inward and outward. In the inward pass, each node send a message to its inward neighbor, and in the outward pass, each node sends a message to its outward neighbors.

In the Hugin architecture, in the inward pass the sender does not divide the message. Instead, we save it in the separator. This requires more space, but it saves computations (as we will see shortly). On the outward pass, the separator divides the outward message by the message it has stored before passing it on to be multiplied into the potential of the receiving node. Notice that the division is done in the separator, which has a smaller state space than either of the two cliques.

If we assume that at the beginning, each separator has the corresponding identity potential ι (a potential whose val-ues are identically one, and whose domain is same as the separator), then the inward action is same as the outward. Precise rules are as follows [Shafer 1996]:

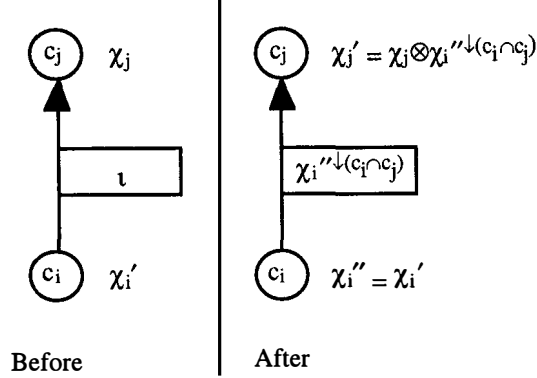

Figure 5. Inward Propagation (from $c_i$ to $c_j$) in the Hugin Architecture

Before | After

- **Rule 1**. Each non-root node waits to send its message to a given neighbor until it has received messages from all its other neighbors.
- **Rule 2**. The root waits to send messages to its neighbors until it has received messages from them all.
- **Rule 3**. When a node is ready to send its message to a particular neighbor, it computes the message by marginalizing its current potential to its intersection with this neighbor, and then it sends the message to the separator between it and the neighbor.
- **Rule 4**. When a separator receives a message *New* from one of its two nodes, it divides the message by its current potential *Old*, sends the quotient *New/Old* on to the other node, and then replaces *Old* with *New*.
- **Rule 5**. When a node receives a message, it replaces its current potential with the product of the potential and the message.

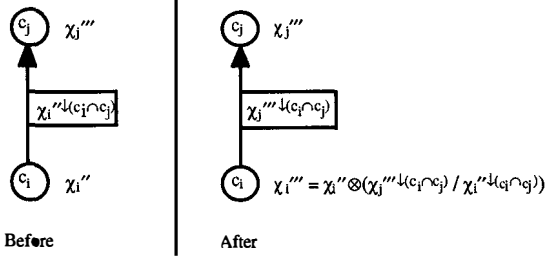

Figure 6. Outward Propagation (from $c_j$ to $c_i$) in the Hugin Architecture

Before | After

Rules 1 and 2 force the propagation to move in to the root and then back out.

At the end of the propagation, the potentials on all the nodes and separators are marginals of the posterior joint P $\propto \otimes \chi_i$ where $\chi_i$ denotes the initial potentials at the clique nodes.

Suppose $\Theta$ is the set of all cliques, and $\Gamma$ is the set of all separators. Then at the beginning, at the end of the inward pass, at the end of the outward pass, or at any step in the propagation process, $P \propto (\otimes_{i \in \Theta} \chi_i) / (\otimes_{i \in \Gamma} \chi_i)$.

We compute the marginal for a variable from the marginal for a smallest separator that contains the variable. If there is no separator that contains the variable then we compute it from the marginal for the clique that contains the variable. Like in the LS architecture, this strategy ignores the computational cost of identifying the smallest separator or clique that contains the variable.

## 5 SHENOY-SHAFER ARCHITECTURE

In this section, we sketch the Shenoy-Shafer architecture and illustrate it using the Chest Clinic problem.

In the Shenoy-Shafer architecture, we start with a collection of potentials that define the joint distribution. The domains of the potentials form a hypergraph. In the Chest Clinic problem, we start with a set of potentials $\vartheta = \{\alpha, \sigma, \tau, \lambda, \beta, \varepsilon, \xi, \delta, o_A, o_D\}$, and a hypergraph $H = \{\{A\}, \{S\}, \{A, T\}, \{S, L\}, \{S, B\}, \{T, L, E\}, \{E, X\}, \{E, B, D\}, \{D\}\}$.

The first step in the Shenoy-Shafer architecture is to arrange the subsets in H in a binary join tree. A binary join tree is a join tree such that no node has more than three neighbors. The binary join tree construction process is motivated by the idea of fusion [Shenoy 1992] (also called peeling by Cannings *et al.* [1978]), and the idea that all combinations should be done on a binary basis, i.e., we multiply potentials two at a time. A binary join tree is a data structure designed to cache computation to reduce the computation involved in combination and marginalization. A binary join tree for the hypergraph in the Chest Clinic problem is shown in Figure 7.

Shenoy [1997] describes a formal procedure for constructing a binary join tree. In general, this binary join tree will have many duplicate subsets. If we have a pair of duplicate nodes that are neighbors and merging these two nodes does not increase the number of neighbors of the merged node to more than three, then we can merge the duplicated nodes into one node. If we do this, we get a binary join tree that we call the condensed binary join tree. In general, we may not be able to always get rid of duplicate nodes [Shenoy 1997].

Notice that the condensed binary join tree obtained may not have all singleton subsets. In the SS architecture, we can compute marginals for any subset in the join tree. We will assume that one is interested in computing marginals for all singleton subsets. So it is important to have all singleton subsets in the binary join tree. In the last step, we attach singleton subsets to the condensed binary join tree. A singleton subset can be included in a binary join tree in many ways. In [Lepar and Shenoy 1997], we describe a method designed to minimize the number of computations required to compute the marginal for that singleton subset. In short, we find a smallest subset that contains the variable we are trying to attach. If there are several such subsets, we select one that minimizes the number of computations required to compute the marginal for that node and we attach the singleton subset to it. Finally, if attaching a singleton subset make a binary tree non-binary, we make the tree binary again by making two copies of the node that has four neighbors, distributing the four neighbors equally between the two copies (two each), and then putting an edge between the two copies. In the Chest Clinic example, our procedure gives us a binary join tree as shown in Figure 7.

**Figure 7.** A Binary Join Tree for the Chest Clinic Problem

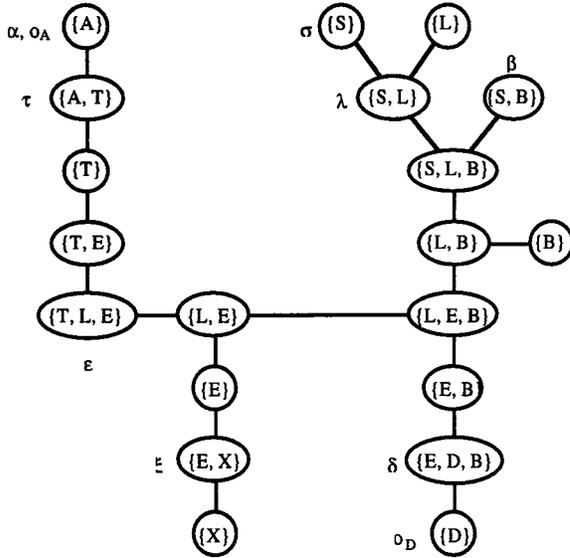

Once we have a binary join tree that contains all singleton subsets, we associate each potential with one of the subsets in the binary join tree that corresponds to its domain. Next, each node in the tree that needs to compute the marginal for it requests a message from each of its neighbors. The messages are computed using Rule 1 as follows.

**Rule 1** (*Computing Messages*) Suppose $r$ and $s$ are neighbors, and suppose $s$ has requested a message from $r$. $r$ in turn requests messages from its other neighbors, and after it has received these messages, it computes the message to $s$ as follows. Informally, the message that node $r$ sends to its neighbor $s$ is the combination of all messages that $r$ receives from its other neighbors together with its own potential marginalized to $r \cap s$. Formally, suppose $\mu^{r \to s}$ denotes the message from $r$ to $s$, suppose $N(r)$ denotes the neighbors of $r$ in the binary join tree, and suppose $\alpha_r$ denotes the probability potential associated with node $r$. Then the message from node $r$ to its neighboring node $s$ is computed as follows:

$$\mu^{r \to s} = (\otimes \{\mu^{t \to r} \mid t \in (N(r) \setminus \{s\})\} \otimes \alpha_r)^{\downarrow r \cap s}$$

Notice that a leaf of the join tree has only one neighbor and therefore when it has received a request for a message, it can send it right away without waiting for any messages.

When a node that needs to compute the marginal for it has requested and received messages from all its neighbors, then it computes the desired marginal using Rule 2 as follows.

**Rule 2** (*Computing Marginals*) When a node $r$ has received a message from each of its neighbors, it combines all messages together with its own probability potential and reports the results as its marginal. If $\varphi$ denotes the joint potential, then

$$\varphi^{\downarrow r} = \otimes \{\mu^{t \to r} \mid t \in N(r)\} \otimes \alpha_r$$

**Figure 8.** The Shenoy-Shafer Architecture for a Join Tree with Two Nodes

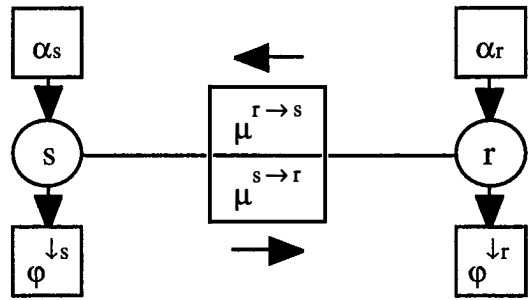

Each node in the binary join tree will have zero, one, two or more storage registers, one for each input probability potential (if any), and one for reporting the marginal of the joint (if a marginal for the node is desired). Each edge (separator) in the join tree would have at most two storage registers for the two messages, one in each direction. Figure 8 shows the storage architecture for a simple join tree with two nodes. Each of the two nodes is assumed to have one input potential. Also, we assume that we desire the marginal for both nodes. Notice that the domain of the separator between $r$ and $s$ is $r \cap s$.

In the Chest Clinic problem, suppose we desire marginals for each of the variables in the problem. To achieve this, suppose that the singleton nodes {A}, {S}, {T}, {L}, {B}, {E}, {X}, and {D} in the binary join tree of Figure 7 request a message from their neighbors. Notice that not all messages are computed. For example, the message $\mu^{SLB \to SB}$ is not computed since it is not requested by any node.

Notice that unlike the LS and Hugin architectures, there are no division operations in the SS architecture. Also, notice that unlike the LS and Hugin architectures, the input potentials remain unchanged during the propagation process in the SS architecture. Notice also that the marginal of the joint potential for a variable is computed at the corresponding singleton variable node of the binary join tree.

## 6 COMPARISON

In this section, we will compare the three architectures. We will focus our attention on the graphical structure for message propagation, the message-passing scheme, the computational efficiency and the storage efficiency of each architecture.

In all three architectures, we assume that we start with a Bayesian network representation of a problem and that we have some evidence (observations or likelihoods) for some variables. The task is to compute the marginals of the posterior distribution for all variables in the problem.

**Graphical Structures for Message Propagation.** In the LS and Hugin architectures, propagation of potentials is done in a junction tree. In the SS architecture, propagation of potentials is done in a binary join tree. The nodes of a junction tree are the cliques of a triangulated moral graph of the original Bayesian network. A corresponding binary join tree includes these cliques as well as several subsets of these cliques. Therefore, a binary join tree has more nodes than in a corresponding junction tree. For example, in the Chest Clinic problem, the junction tree shown in Figure 2 has six nodes whereas the corresponding binary join tree shown in Figure 7 has 20 nodes. (Notice that if we start with a binary join tree and we condense it by absorbing adjacent nodes that are subsets/supersets of each other, we get what we call a "corresponding junction tree.")

The junction tree yields only marginals for the cliques in the LS architecture, and marginals for cliques and separators in the Hugin architecture. Since our stated task is to compute marginals of all singleton variables, there is further computation needed in these two architectures. In the LS architecture, the marginal for a variable can be computed most efficiently from the marginal of a smallest clique containing the variable. In the Hugin architecture, if a variable belongs to a separator, then the marginal for the variable can be computed most efficiently from a smallest separator containing the variable. If a variable does not belong to any separator, then its marginal can be computed most efficiently from the clique containing the variable. In the SS architecture, since we include all singleton subsets during the construction of a binary join tree, the graphical structure yields marginals for singletons at the end of the message passing stage.

It is not necessary that we use junction trees for the LS and Hugin architectures. We could use any join tree including binary join trees. However, given the message passing schemes of these two architectures, it is inefficient (with respect to both computation and storage) to implement these two message-passing schemes on join trees with many nodes. We will be more specific about this aspect when we discuss computational efficiencies of the three architectures. Also, it is not necessary that we use a binary join tree for the SS architecture. We could use any join tree including junction trees. However, there is computational penalty in using non-binary join trees or junction trees for the SS message-passing scheme. For these reasons, the LS architecture is associated with junction trees, the Hugin architecture is associated with junction tree with separators, and the SS architecture is associated with binary join trees constructed in the manner described in [Lepar and Shenoy 1997].

**Message-Passing Schemes.** In the LS architecture, first we designate a maximal clique (with the largest state space) of the junction tree as the root. The propagation of messages in done in two stages—the inward phase where each clique send a message to its inward neighbor, and the outward phase in which each clique sends a message to each of its outward neighbors. At the beginning, we have an evidence potential representation. And at the end of the outward phase, at each clique, we have the marginals for it. Each clique in the junction tree stores a potential. Computations are done by each clique in the junction tree.

In the Hugin architecture, we designate a node as the root. Each clique sends a message to each of the separators between it and its neighbors. When a separator receives a message from one of its neighboring clique, it sends a message to its other neighboring clique. At all times the joint potential is equal to the product of the potentials at the cliques divided by the product of the potentials at the separators. When all messages have been sent, the potential at each clique and at each separator is the marginal of the joint for that node. Each clique and each separator in the junction tree stores a potential. Computations are done by each clique and by each separator in the junction tree.

In the SS architecture, nodes for which the marginals are desired request messages from all their neighbors. When a node receives a request for a message, it in turn requests messages from all its other neighbors. When all requested messages have been delivered, the marginals are computed at the desired nodes. A node may store either no potential, or one potential (input or output) or two or more potentials (one for each input and output). Each edge (separator) between two nodes may store one or two potentials. Computations are done only by nodes and not by separators.

Although we have restricted our study in this article to Bayesian networks, all three architectures are applicable more widely. Lauritzen and Jensen (1996) have described axioms that generalize the LS and the Hugin architecture to other domains. These axioms include the axioms proposed by Shenoy and Shafer (1990). A natural question is how generally applicable are these three architectures. Since the Shenoy-Shafer architecture does not use the division operation, it is clear that the Shenoy-Shafer architecture is more widely applicable that the Lauritzen-Spiegelhalter or the Hugin architecture. For example, the problem of fast retraction proposed by Cowell and David [1992] can be handled by all three architectures in the probabilistic domain. However, fast retraction cannot be handled in non-probabilistic domains by the Lauritzen-Spiegelhalter and Hugin architectures, as the axioms are not satisfied [Lauritzen and Jensen 1996]. Fast retraction is easily handled in the Shenoy-Shafer architecture [Lauritzen and Shenoy 1996]. Another example is a system of a finite number of logical formulae with a definition of combination and marginalization [Kohlas and Stärk 1996]. The SS architecture can then be used to compute the marginals [Lepar 1998]. Since there is no natural definition of the division operation in this system, neither the LS nor the Hugin architecture can be used to compute marginals.

Table 1. # Binary Arithmetic Operations for Ten Sample Problems

| # Binary Arithmetic Operations<br><br>Random Bayes Net (with $c_1 = 5$) | Architecture | | |
|---|---|---|---|
| | LS<br>(+, ×, ÷) | Hugin<br>(+, ×, ÷) | SS<br>(+, ×, ÷) |
| #1 with 8 variables, m = 6, p = 3, $c_2 = 2$<br>TOTAL # operations | (179, 202, 49)<br>430 | (161, 202, 16)<br>379 | (129, 200, -)<br>329 |
| #2 with 8 variables, m = 2, p = 3, $c_2 = 2$<br>TOTAL # operations | (52, 64, 20)<br>136 | (46, 64, 10)<br>120 | (34, 78, -)<br>112 |
| #3 with 8 variables, m = 2, p = 3, $c_2 = 2$<br>TOTAL # operations | (48, 68, 24)<br>140 | (40, 68, 10)<br>118 | (34, 86, -)<br>120 |
| #4 with 8 variables, m = 2, p = 3, $c_2 = 4$<br>TOTAL # operations | (176, 192, 64)<br>432 | (136, 192, 32)<br>360 | (132, 268, -)<br>400 |
| #5 with 10 variables, m = 8, p = 3, $c_2 = 2$<br>TOTAL # operations | (507, 455, 188)<br>1,150 | (468, 455, 34)<br>957 | (397, 441, -)<br>838 |
| #6 with 10 variables, m = 2, p = 3, $c_2 = 2$<br>TOTAL # operations | (60, 76, 28)<br>164 | (52, 76, 14)<br>142 | (42, 98, -)<br>140 |
| #7 with 12 variables, m = 4, p = 3, $c_2 = 2$<br>TOTAL # operations | (361, 348, 120)<br>829 | (324, 348, 35)<br>707 | (248, 354, -)<br>602 |
| #8 with 12 variables, m = 2, p = 3, $c_2 = 2$<br>TOTAL # operations | (72, 96, 36)<br>204 | (58, 96, 18)<br>172 | (50, 116, -)<br>166 |
| #9 with 12 variables, m = 2, p = 3, $c_2 = 5$<br>TOTAL # operations | (1152, 1365, 222)<br>2,739 | (867, 1365, 66)<br>2,298 | (783, 1419, -)<br>2,201 |
| #10 with 16 variables, m = 4, p = 5, $c_2 = 2$<br>TOTAL # operations | (590, 587, 177)<br>1,354 | (526, 587, 41)<br>1,154 | (570, 412, -)<br>982 |

Hugin architectures and binary join trees for the SS architecture making sure that the maximal subsets (cliques) are the same in all join trees generated. Finally, we count binary arithmetic operations required by each architecture for computing marginals for each variable in the network. Table 1 shows the number of binary arithmetic operations required by the three architectures for ten randomly generated Bayes nets. The corresponding join trees for the ten Bayes nets and the formulae for counting the operations are in Lepar [1998].

**Computational Efficiencies.** It is traditional to study worst case order of magnitude complexity of computational algorithms. From this perspective, there are no essential differences between the three architectures. All three architectures compute the marginals using local computation. In the worst case, the computational complexity of the three algorithms are exponential in the size (# variables) of the maximal clique.

Here we will look at computational efficiencies of the three architectures using a very crude measure: # binary arithmetic operations (additions, multiplications, and divisions). It is clear that this crude measure does not describe the actual computational efficiency. This measure does not include other operations such as table lookups, comparisons, read/write to memory, construction of the graphical data structure, etc. Our methodology is as follows. First we randomly generate Bayesian networks using some controlling parameters such as number of variables (n), the maximum distance of a node from its neighbors in the ordering consistent with the arrows in a Bayes net ($c_1$), the maximum number of parents (or children) with which a variable is added to the Bayes net ($c_2 - 1$), the maximum state space for variables (m), and the maximum number of variables for which we have observations (p). The number of parents (or children) of a node, with which it is added to the Bayes net, is selected randomly from the interval [1, $c_2$). The size of state space of each variable in the network is selected randomly from the interval [2, m]. The number of variables for which we have evidence (observations or likelihoods) is chosen randomly from the interval [1, p]. The algorithm used to generate random Bayes nets and the inequality for the total number of neighbors are in Lepar [1998]. Next, we generate junction trees for the LS and

First, the Hugin architecture always does fewer additions than the LS architecture. This is because computation of marginals of singleton variables is always done from clique marginals in the LS architecture whereas in the Hugin architecture, it is done from the separator marginals for some variables and clique marginals for some variables. Notice that we are not including the computational cost of identifying a smallest clique in the LS architecture and the cost of finding a smallest separator or clique in the Hugin architecture (since this is part of the process of compiling a Bayes net to a junction tree, and there is a similar optimization in the SS architecture in adding singleton subsets to the condensed binary join tree).

Second, the LS and Hugin architectures always do the same number of multiplications. The Hugin architecture is an adaptation of the LS architecture, and it is not surprising that this aspect of the two architectures is the same.

Third, the Hugin architecture always does fewer divisions than the LS architecture. The Hugin architecture does divisions in the separator whereas the LS architecture does divisions in the cliques. This major motivation led to the Hugin architecture. Since the Hugin architecture is always more computationally efficient than the LS architecture, we will restrict our comparison of the SS architecture to the Hugin architecture.

Comparing the Hugin and SS architectures, in Table 1, we notice that for most of the Bayes nets generated, the total number of operations in the SS architecture is less than in the Hugin architecture. Given the number of nodes, the total number of operations depends strongly on the parameter m (maximum size of the state space for a variable), and the parameter $c_2$ which influences the sizes of cliques. To fully compare the two architectures in the average case, we generated 20,000 random Bayes nets with a fixed set of parameters and computed the average total number of operations for the three architectures. We did this for several combinations of the parameters. The results are shown in Table 2. On an average, SS uses fewer operations than Hugin. When we have variables with larger state spaces, the difference between Hugin and SS is greater. When m = 3, Hugin requires approximately 6% more computations than SS. When m = 6, Hugin requires approximately 40% more. These results contradict the popular perception that Hugin is the most efficient architecture for computing marginals.

In Tables 1 and 2, we have been assuming that one division is equal to one addition, and one division is equal to one multiplication. This is not true for most processors. For example, for the Intel Pentium MMX 200Mhz microprocessor, one division (double precision) takes approximately as much time as three multiplications. For the Motorola PowerPC 604e 132Mhz microprocessor, one division (double precision) takes approximately as much time as 10 multiplications. On the Sun SPARC processor, one division (double precision) takes approximately as much time as 2.3 multiplications. Since the SS architecture does not do divisions, these factors add to the computational efficiency of the SS architecture.

Since there is no guarantee of doing binary multiplications in a junction tree, the Hugin architecture does more multiplications than is done by the SS architecture in a binary join tree (where all multiplications are done on a binary basis).

Notice that the Hugin propagation can be done in any join tree assuming we start with a clique marginal representation of the joint probability distribution. However since computations are done at each node and at each separator, there is a computational penalty in introducing additional nodes and separators. For example for the Chest Clinic problem with evidence for A and D, if we do Hugin propagation in the binary join tree shown in Figure 7, it requires 60 additions, 116 multiplications and 46 divisions for a total of 222 operations, compared to 60 additions, 96 multiplications and 16 divisions for a total of 172 operations [for details see [Lepar and Shenoy 1997]). Clearly, for the Hugin architecture, the junction tree in Figure 2 is more efficient than the binary join tree of Figure 7. Can one improve on the junction tree data structure for the LS and Hugin architectures? Almond and Kong [1991] and Kjærulff [1997] suggest new data structures. We have yet to study the nested junction tree data structure suggested recently by Kjærulff. The join trees suggested by Almond and Kong [1991] are less efficient for the LS and Hugin architectures than the junction trees studied here. They introduce a "separator" node between two or more neighboring nodes. The Hugin architecture is only defined for a separator node between exactly two neighboring nodes. We can of course consider all "separator" nodes in the Almond-Kong join trees as ordinary nodes. But, in this case, we get a join tree similar to the binary join trees considered here (in the sense that the join tree has much more nodes than in the junction tree) and we have seen that this is less efficient for Hugin architecture than junction trees.

The SS propagation can be done in any join tree assuming we start with an evidence potential representation of the joint probability distribution. Since no computations are done at any separator, and since the computations done at a node depends on the number of neighbors, there may be a computational penalty if we use arbitrary join trees. For example for the Chest Clinic problem with evidence for A and D, if we do SS propagation in the junction tree shown in Figure 2, it requires 60 additions, and 140 multiplications for a total of 200 operations compared to 56 additions and 124 multiplications for a total of 180 operations for the binary join tree of Figure 7 (for details see Lepar and Shenoy 1997]). Clearly, for the SS architecture, the binary join tree in Figure 7 is more efficient than the junction tree of Figure 2 even though the junction tree of Figure 2 is a binary join tree. Notice that if we restrict ourselves to cliques, there is no guarantee that we can always find a junction tree that is a binary join tree. Other join tree construction procedures have been proposed for the SS architecture (see, e.g., [Almond 1995, pp. 157–168]), but there have been no systematic study of the efficiency of these methods. We conjecture that the binary join tree construction process as described in this article always produces join trees which are more efficient for the SS architecture than those produced by the Almond-Kong procedure (which does not guarantee to produce binary join trees).

Table 2. Average Case Analysis of the Computational Efficiencies of the Three Architectures

| Random Bayes nets (with p = 3) (m = max. size of state space of variables) | Average total # binary operations (from 20,000 random cases) | | |
|---|---|---|---|
| | LS | Hugin | SS |
| 6 variables, m = 3, $c_2$ = 2 | 107 | 96 | 93 |
| 6 variables, m = 6, $c_2$ = 4 | 2,006 | 1,797 | 1,540 |
| 8 variables, m = 3, $c_2$ = 2 | 169 | 151 | 142 |
| 8 variables, m = 6, $c_2$ = 5 | 11,207 | 10,052 | 8,417 |
| 10 variables, m = 3, $c_2$ = 2 | 216 | 190 | 179 |
| 10 variables, m = 6, $c_2$ = 6 | 63,870 | 57,370 | 48,072 |
| 12 variables, m = 3, $c_2$ = 2 | 265 | 231 | 218 |
| 12 variables, m = 6, $c_2$ = 2 | 1,760 | 1,602 | 1,142 |
| 14 variables, m = 3, $c_2$ = 8 | 2,597,490 | 2,368,992 | 2,082,052 |
| 14 variables, m = 6, $c_2$ = 2 | 2,042 | 1,841 | 1,337 |
| 16 variables, m = 3, $c_2$ = 9 | 34,609 | 31,035 | 30,491 |
| 16 variables, m = 6, $c_2$ = 2 | 2,424 | 2,174 | 1,584 |
| 20 variables, m = 3, $c_2$ = 2 | 476 | 407 | 387 |
| 20 variables, m = 6, $c_2$ = 2 | 3,190 | 2,833 | 2,097 |
| 25 variables, m = 3, $c_2$ = 2 | 616 | 523 | 500 |
| 25 variables, m = 6, $c_2$ = 3 | 982,944 | 924,656 | 813,458 |
| 30 variables, m = 3, $c_2$ = 2 | 137,390 | 128,442 | 117,138 |
| 41 variables, m = 3, $c_2$ = 2 | 2,512,694 | 2,371,178 | 2,072,072 |

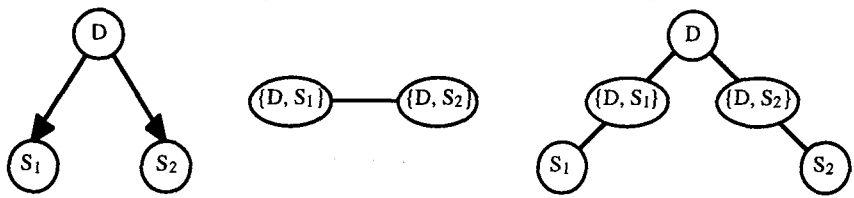

Figure 9. A Bayes Net, a Junction Tree, and a Binary Join Tree

**Storage Efficiencies.** In the LS architecture, each clique in the junction tree stores one potential. Thus, the total storage requirements will depend on the number of cliques in the junction tree and state spaces of the cliques. If after propagating the messages in the junction trees, we get a new piece of evidence, then we will have to start again with the input and evidence potentials. Also, a user may want to edit the input and evidence potentials. For these two reasons, we have to also include the storage requirements for the input and evidence potentials. Also, at the end of the outward propagation, we have only the marginals for the cliques. However, our stated task is the computation of the marginals for each variable. These marginals are computed from the clique marginals. We will also include the storage requirements for storing the marginals of each variable.

In the Hugin architecture, each clique in the junction tree stores one potential. Also, each separator between two adjacent cliques stores one potential. Also, a user may need to edit the input and evidence potentials. Therefore, these need to be stored separately. Therefore, we will also include the storage space for storing the input and evidence potentials. Also, when all messages have been computed, we have only the marginals for the cliques and separators. We still need to compute marginals of singleton variables. Therefore, we will include storage space for the marginals of each variable.

In the SS architecture, each node may have either zero, one, two or more potentials. If a node has at least one input potential and it is a singleton node whose marginal is desired, then such a node will have two or more potentials. If a node has neither an input potential nor is the marginal for the node desired, then it will have zero potentials. In all other cases, it will have one potential (either an input potential or an output potential). If we regard the edge between two adjacent nodes as a separator, then each separator will have either one or two potentials depending on which messages are requested. If both adjacent nodes request messages from each other, then each separator will store two potentials. If only one message is requested, then a separator will store only one potential.

In general, it is easy to see that, assuming we are working with the same junction tree, the Hugin architecture will have always more storage requirements than the LS architecture because of storage at the separators.

In comparing the storage requirements of Hugin with SS architectures, there are no general results. Although a binary join tree has more nodes than a corresponding junction tree, not every node in a binary join tree has a potential associated with it. All input and evidence potential are included in both architectures and all output potentials are also included in both architectures. So the differences in storage are due to storage at cliques and separators in the Hugin architecture and storage at separators in the SS architecture. In the Hugin architecture, all separators include exactly one potential each, whereas in the SS architecture, most separators include two potentials and there are usually a lot more separators in a binary join trees than in corresponding junction trees. However, every clique in a junction tree stores a potential whereas these potentials are not present in the SS architecture.

For the Chest Clinic example, the SS architecture has more storage than the LS and Hugin architectures. It is easy to construct an artificial problem in which the SS architecture has less storage than the LS and Hugin architectures. Consider a Bayes net with one disease variable D and two symptom variables $S_1$ and $S_2$ as shown in Figure 9. Suppose we have two pieces of evidence for nodes $S_1$ and $S_2$, respectively. A junction tree and a binary join tree are also shown in Figure 9. Suppose that each of the three variables has 5 states. Then in all three architectures we have the same storage for input (5 + 25 + 25 = 55 fpn), evidence (5 + 5 = 10 fpn) and output potentials (3*5 = 15 fpn). In the LS architecture we have a storage of 50 (= 2*25) fpn at the two cliques in the junction tree. In the Hugin architecture, we have a total storage of 55 (= 2*25 + 5) fpn at the two cliques and one separator. In the SS architecture, we have a total storage of 40 (= 4*2*5) at the 4 separators. Thus in this problem, SS has less storage than both the LS and Hugin architectures. Our storage analysis doesn't take into account the runtime working memory required for computing the messages. In the SS architecture, before we do the marginalizing operation, we have to combine all messages on the state space of the subset node. Thus for the SS architecture, we would need to add the size of the state space of the largest subset in the join tree to our count.

More study is required to compare the storage efficiencies of Hugin and SS architectures. This task remains to be done. We conjecture that if we study the storage requirements for randomly generated Bayes nets, on an average, the Hugin architecture will be more storage efficient than the SS architecture.

## 7 CONCLUSIONS

Overall, comparing LS and Hugin architectures, Hugin is computational more efficient than LS, whereas LS is more storage efficient than Hugin. In a sense, Hugin sacrifices storage efficiency to achieve better computational efficiency. On an average, SS is computationally more efficient than Hugin. We conjecture that on an average, Hugin is more storage efficient than SS. This task remains to be done.

Our study here also ignores the dynamics aspects of computing marginals. For example, suppose we have already computed the marginals for all singleton subsets, and we get a new item of evidence. It is not necessary to re-compute all the messages. In all three architectures, we can compute the revised marginals by re-computing some of the messages. A detailed study of the efficiency of updating the marginals is another task that remains to be done.

Since we started our project, Kjærulff [1997] has proposed a new data structure called nested junction trees for the Hugin and SS architectures, and Schmidt and Shenoy [1998] have proposed some improvements for the SS and Hugin architectures. The results presented here do not take into account fully some of these new proposed changes. This is yet another task that remains to be done.

### Acknowledgments

The authors are grateful for support and encouragement from Professor Jürg Kohlas. We are especially grateful to Robert Stärk and Norbert Lehmann for their help on this project. The paper has benefited also from comments and suggestions by Bernard Anrig, Roman Bissig, Rolf Haenni, Jürg Kohlas, Yves Kreis, Paul-Andre Monney, Dennis Nilsson, Tuija Schmidt, and three anonymous referees.

### References

Almond, R. (1995), *Graphical Belief Modeling*, Chapman & Hall, London.

Almond, R. and A. Kong (1991), "Optimality issues in constructing a Markov tree from graphical models," Research Report No. 329, Department of Statistics, University of Chicago, Chicago, IL.

Cannings, C., E. A. Thompson and M. H. Skolnick (1978), "Probability functions on complex pedigrees," *Advances in Applied Probability*, 10, 26–61.

Cowell, R. and A. P. Dawid (1992), "Fast retraction of evidence in a probabilistic expert system," *Statistics and Computing*, 2, 37–40.

Jensen, F. V., K. G. Olesen and S. K. Andersen (1990a), "An algebra of Bayesian belief universes for knowledge-based systems," *Networks*, 20(5), 637–659.

Jensen, F. V., S. L. Lauritzen and K. G. Olesen (1990b), "Bayesian updating in causal probabilistic networks by local computation," *Computational Statistics Quarterly*, 4, 269–282.

Kjærulff, U. (1997), "Nested junction trees," in D. Geiger and P. P. Shenoy (eds.), *Uncertainty in Artificial Intelligence: Proceedings of the Thirteenth Conference*, 294–301, Morgan Kaufmann, San Francisco, CA.

Kohlas, J. and R. Stärk (1996), "Information algebras and information systems," Working Paper No. 96-12, Institute of Informatics, University of Fribourg, Fribourg, Switzerland.

Lauritzen, S. L. and D. J. Spiegelhalter (1988), "Local computations with probabilities on graphical structures and their application to expert systems (with discussion)," *Journal of Royal Statistical Society, Series B*, 50(2), 157–224.

Lauritzen, S. L. and F. V. Jensen (1996), "Local computation with valuations from a commutative semigroup," Technical Report No. R-96-2028, Institute for Electronic Systems, Department of Mathematics and Computer Science, Aalborg University, Aalborg, Denmark.

Lepar, V. (1998), "Performance of Architectures for Local Computations in Bayesian Networks," PhD dissertation, Institute of Informatics, University of Fribourg, Fribourg, Switzerland, in preparation.

Lepar, V. and P. P. Shenoy (1997), "A Comparison of Architectures for Exact Computation of Marginals," Working Paper No. 274, School of Business, University of Kansas, Lawrence, KS.

Lauritzen, S. L. and P. P. Shenoy (1996) "Computing marginals using local computation", Working Paper No 267, School of Business, University of Kansas, Lawrence, KS.

Pearl, J. (1986), "Fusion, propagation and structuring in belief networks," *Artificial Intelligence*, 29, 241–288.

Schmidt, T. and P. P Shenoy (1998), "Some improvements to the Shenoy-Shafer and Hugin architectures for computing marginals," *Artificial Intelligence*, in press.

Shafer, G. (1996), *Probabilistic Expert Systems*, Society for Industrial and Applied Mathematics, Philadelphia, PA.

Shenoy, P. P. (1992), "Valuation-based systems: A framework for managing uncertainty in expert systems," in L. A. Zadeh and J. Kacprzyk (eds.), *Fuzzy Logic for the Management of Uncertainty*, 83–104, John Wiley & Sons, New York, NY.

Shenoy, P. P. (1997), "Binary join trees for computing marginals in the Shenoy-Shafer architecture," *International Journal of Approximate Reasoning*, 17(1), 1–25.

Shenoy, P. P. and G. Shafer (1986), "Propagating belief functions using local computation," *IEEE Expert*, 1(3), 43–52.

Shenoy, P. P. and G. Shafer (1990), "Axioms for probability and belief-function propagation," in R. D. Shachter, T. S. Levitt, J. F. Lemmer and L. N. Kanal (eds.), *Uncertainty in Artificial Intelligence, 4*, 169–198, North-Holland, Amsterdam.